\documentclass[pmlr]{jmlr}

\usepackage{longtable}
\usepackage{graphicx}
\usepackage{booktabs}
\usepackage[load-configurations=version-1]{siunitx} 
\usepackage{natbib}
\usepackage{bm}

\graphicspath{ {./images/} }
\usepackage{caption}
\usepackage{subcaption}

\newcommand{\Amat}[0]{\ensuremath{{\bf A}}}

\newcommand{\Kmat}[0]{\ensuremath{{\bf K}} }

\newcommand{\Omat}[0]{\ensuremath{{\bf O}} }

\newcommand{\Qmat}[0]{\ensuremath{{\bf Q}} }

\newcommand{\Smat}[0]{\ensuremath{{\bf S}} }

\newcommand{\Vmat}[0]{\ensuremath{{\bf V}} }
\newcommand{\Wmat}[0]{\ensuremath{{\bf W}} }
\newcommand{\Xmat}[0]{\ensuremath{{\bf X}} }

\newcommand{\bv}[0]{\ensuremath{\boldsymbol{b}} }

\newcommand{\ev}[0]{\ensuremath{\boldsymbol{e}} }

\newcommand{\hv}[0]{\ensuremath{\boldsymbol{h}} }

\newcommand{\kv}[0]{\ensuremath{\boldsymbol{k}} }

\newcommand{\ov}[0]{\ensuremath{\boldsymbol{o}} }

\newcommand{\qv}[0]{\ensuremath{\boldsymbol{q}} }

\newcommand{\tv}[0]{\ensuremath{\boldsymbol{t}} }
\newcommand{\uv}[0]{\ensuremath{\boldsymbol{u}} }
\newcommand{\vv}[0]{\ensuremath{\boldsymbol{v}} }

\newcommand{\xv}[0]{\ensuremath{\boldsymbol{x}} }

\theorembodyfont{\upshape}
\theoremheaderfont{\scshape}
\theorempostheader{:}
\theoremsep{\newline}

\jmlrvolume{}
\jmlryear{2020}
\jmlrworkshop{Machine Learning for Healthcare}

\title[BERT-based Attention Model for Message Triage]{Students Need More Attention: BERT-based Attention Model for Small Data with Application to Automatic Patient Message Triage}

\author{\Name{Shijing Si}
      \Email{shijing.si@duke.edu}\\ 
      \addr Department of Electrical and Computer Engineering\\
      Duke University\\
      Durham, NC, USA 
      \AND
      \Name{Rui Wang}
      \Email{rw161@duke.edu}\\ 
      \addr Department of Electrical and Computer Engineering\\
      Duke University\\
      Durham, NC, USA 
      \AND
      \Name{Jedrek Wosik}
      \Email{jedrek.wosik@duke.edu}\\ 
      \addr Department of Electrical and Computer Engineering\\
      Duke University\\
      Durham, NC, USA 
      \AND
      \Name{Hao Zhang}
      \Email{hz210@duke.edu}\\ 
      \addr Department of Electrical and Computer Engineering\\
      Duke University\\
      Durham, NC, USA 
      \AND
      \Name{David Dov}
      \Email{david.dov@duke.edu}\\ 
      \addr Department of Electrical and Computer Engineering\\
      Duke University\\
      Durham, NC, USA 
      \AND
      \Name{Guoyin Wang}
      \Email{guoyin.wang@duke.edu}\\ 
      \addr Department of Electrical and Computer Engineering\\
      Duke University\\
      Durham, NC, USA 
      \AND
      \Name{Ricardo Henao}
      \Email{ricardo.henao@duke.edu}\\ 
      \addr Department of Electrical and Computer Engineering\\
      Duke University\\
      Durham, NC, USA 
      \AND
      \Name{Lawrence Carin}
      \Email{lcarin@duke.edu}\\ 
      \addr Department of Electrical and Computer Engineering\\
      Duke University\\
      Durham, NC, USA 
      } 

\editor{Editor's name}

\begin{document}

\maketitle

\begin{abstract}
Small and imbalanced datasets commonly seen in healthcare represent a challenge when training classifiers based on deep learning models.
So motivated, we propose a novel framework based on BioBERT (Bidirectional Encoder Representations from  Transformers for Biomedical TextMining).
Specifically, ($i$) we introduce Label Embeddings for Self-Attention in each layer of BERT, which we call LESA-BERT, and ($ii$) by distilling LESA-BERT to smaller variants, we aim to reduce overfitting and model size when working on small datasets.
As an application, our framework is utilized to build a model for patient portal message triage that classifies the urgency of a message into three categories: non-urgent, medium and urgent.
Experiments demonstrate that our approach can outperform several strong baseline classifiers by a significant margin of 4.3\% in terms of macro F1 score.  The code for this project is publicly available at \url{https://github.com/shijing001/text_classifiers}
\end{abstract}

\section{Introduction}
Online patients portals, \emph{e.g.}, MyChart by Epic, have become increasingly prevalent tools for communication between patients and healthcare providers \citep{ramsey2018increasing}. 
These portals have the potential to boost the productivity of providers, improve patient satisfaction, and reduce communication barriers \citep{goldzweig2013electronic,sieck2017rules}. 
Despite these benefits, patient-provider communication tools have produced unintended consequences such as increased, often unpaid, workload for providers \citep{hefner2019patient}.
Further, due to the flood of non-urgent incoming patient messages, some, which may require a timely emergency response, can be delayed or effectively neglected.

To address these concerns, we consider the task of automated patient message classification to estimate the urgency of messages based on their content.
Our dataset consists of $1,756$ messages collected from a University Hospital's portal.
These messages, manually adjudicated by experienced healthcare providers, were grouped into three categories: non-urgent, medium, and urgent, as summarized in Table~\ref{tab:data}.
With merely $170$ urgent instances, our dataset is both small and imbalanced, posing a significant challenge in terms of properly training machine-learning-based classifiers.
This challenge is common in many clinical datasets \citep{zhao2018framework,para2019clinical} due to the fact that manually labeling data is very laborious, time-consuming, expensive and oftentimes prohibitive.
Further, certain labels are rare by nature, for instance, urgent electronic messages are far less common because patients tend to call the healthcare provider directly rather than using the portal.

\begin{table}[ht!]
\centering
\begin{tabular}{llp{10.5cm}}
    Label & Count & Typical Example \\ \hline
    Non-urgent    &   631    &     That would be awesome... thank you. \\ \hline
    Medium    &     955  &     Dr. [name]. All seems well now. I am at home resting. My
    wife and I have a trip planned to Maryland this week 
    beginning on Wednesday. We can fly, drive or stay home
    if I should not travel. Are there any reasons that I
    should not fly. \\ \hline
    Urgent    &     170  &       I have continued having chest pain shortness of
    breath since waking. Please tell me what to do. I have tried
    in hailers am going to try nebulizers. I just feel extremely
    tight in my chest. \\
\end{tabular}
\caption{Typical examples of patient messages to providers grouped by urgency. These are examples of the message urgency dataset used in the experiments.}
\label{tab:data}
\end{table}

Machine learning approaches have been previously applied to message classification tasks in the healthcare domain.
For example, \citet{cronin2015automated} utilized logistic regression and random forest algorithms to classify between patient health information needs such as symptom management and medication side effects, based on their messages.
\citet{cronin2017comparison} studied the use of rule-based and random forest classifiers to classify patient portal messages into broad communication types, \emph{e.g.}, appointment rescheduling, examination enquiry, \emph{etc}.
\citet{sulieman2017classifying} showed that Convolutional Neural Networks (CNNs) outperform traditional classifiers in portal message classification,
and \cite{tafti2019artificial} developed an ensemble of neural networks for text classification to categorize free-text patient portal messages as either containing active symptom descriptions or logistic requests.
\citet{chen2019detecting} leveraged traditional machine learning methods such as
Support Vector Machines (SVMs) to detect hypoglycemia incidents reported in patient messages.
To the best of our knowledge, most existing classifiers employ either traditional machine learning methods such as SVMs or shallow networks for the classification of patient generated messages.

Recently, the field of natural language processing (NLP) has seen significant progress in large-scale pre-trained language models, which often considerably boost the performance of classifiers on text classification tasks \citep{devlin2018bert,radford2019language}.
Here, we introduce the use of Bi-directional Encoder Representations from
Transformers (BERT) \citep{devlin2018bert} for patient message classification. 
BERT produces sequence representations using a multi-layered attention mechanism, which models associations between the tokens (words) of a sentence \citep{vaswani2017attention}.
We consider the challenging task of properly training such complex models on a small and imbalanced dataset, which is complicated because these models typically have hundreds of millions of parameters.
Previous studies addressed this challenge by pre-training BERT on a large dataset and then fine-tuning it on the (small) dataset at hand \citep{devlin2018bert}.
However, the performance of this procedure is usually limited by the size and difficulty of the available small dataset.

\subsection*{Generalizable Insights from Machine Learning in the Context of Healthcare}
In this paper we address this challenge with a framework which beyond the use of a pre-trained model, has two novel components.
First, we introduce a novel attention mechanism that utilizes label embeddings to better capture associations 
between the labels and the tokens of the input sequence.
Importantly, the proposed attention mechanism can be incorporated into existing attention layers with minimum modification, allowing the use of pre-trained BERT or BioBert models (\cite{devlin2018bert,lee2020biobert}).
We term our method LESA-BERT, short for Label Embedding on Self-Attention in BERT.
Second, a large deep learning model like BERT has millions of parameters and tends to overfit on small datasets.
We hypothesize that a model with a reduced number of parameters may result in a smaller generalization error, thus likely improved performance.
Accordingly, we employ the knowledge distillation technique to compress the LESA-BERT model, by training a smaller \emph{student} model to reproduce the prediction ability of the \emph{teacher}, a fine-tuned LESA-BERT model.
From our framework, we devise distilled variants of LESA-BERT.
We demonstrate LESA-BERT by building an automatic message triage classifier that can predict the urgency of messages based on their content.
Experiments demonstrate that LESA-BERT and distilled variants result in improved performance compared to multiple baseline approaches. 
Specifically, LESA-BERT outperforms the baselines by a 2.5\% margin in terms of macro F1 score, while the distilled LESA-BERT with 6 encoder layers (Distil-LESA-BERT-6) further outperforms LESA-BERT by 1.8\%.

\section{Related Work}
\paragraph{BERT} Widely used for natural language processing, BERT produces sentence representations through a multi-layered attention mechanism that encodes relations between the tokens that compose the sentence \citep{vaswani2017attention, devlin2018bert}.
Due to its complex structure, which encompasses a large number of parameters, BERT is typically pre-trained on a large dataset and then fine-tuned on the target dataset.
Standard pre-training approaches include unsupervised tasks such as masked language modeling and next sentence prediction, which leverage massive unlabeled, general domain corpora like English Wikipedia (2,500M words) or BooksCorpus (800M words) \citep{zhu2015aligning}.
Subsequently, the model is fine-tuned by adding one or more additional layers, the parameters of which are optimized using the target dataset, in a technique termed transfer learning.
Recently BERT-based models have been leveraged to NLP tasks in healthcare.
\citet{huang2019clinicalbert} developed ClinicalBERT by pre-training
BERT on a large set of clinical notes, and utilized it to predict
patients' 30-day hospital readmission based on both discharge summaries
and clinical notes.
\citet{lee2020biobert} introduced BioBERT, which is essentially BERT pre-trained on large biomedical corpora that includes PubMed abstracts and PubMed Central full-text articles.
They showed that pre-training the model directly on the domain of interest leads to improved performance on various biomedical NLP tasks.

\paragraph{Label Embedding} Label embedding is a technique that embeds class labels along with the (text) data into a joint latent space, where the model can be trained to cross-attend the inputs and labels to boost the performance of deep learning models. 
Label embeddings were previously leveraged for image classification \citep{akata2015label}, multi-modal learning between images and text \citep{kiros2014multimodal}, text recognition in images  \citep{rodriguez2015label}, zero-shot learning \citep{li2015zero,ma2016label} and text classification \citet{zhang2017multi}.
Notably, \citet{wang2018joint} proposed a framework named Label Embedding Attentive Model (LEAM), which jointly embeds the words and labels in a common latent space, and improves the performance on general text classification tasks.
Inspired by LEAM, we consider the joint representation of the message and its corresponding class token, and propose to incorporate label embeddings to the self-attention mechanism inside BERT encoders, which improves the attention between the class token to the other tokens in the message.  

\paragraph{Knowledge Distillation} Knowledge distillation is a technique used to train a small model usually called \emph{student} to reproduce the predictions, thus performance, of a larger model called a \emph{teacher} \citep{hinton2015distilling}.
It is typically used to compress models, which reduces their storage and computational costs, thus facilitating their deployment.
Model distillation has been previously applied to compress complex models such as BERT and LSTMs \citep{sanh2019distilbert, tang2019distilling}.
When the training data size is small, we hypothesize that the teacher model may be prone to overfitting.
In this situation, knowledge distillation is more likely to produce a student model that outperforms the larger, more complex teacher model, as we show in our experiments.

\begin{figure*}[t!]
    \centering
    \begin{subfigure}
        \centering
        \includegraphics[width=0.6\textwidth, height=1.7in]{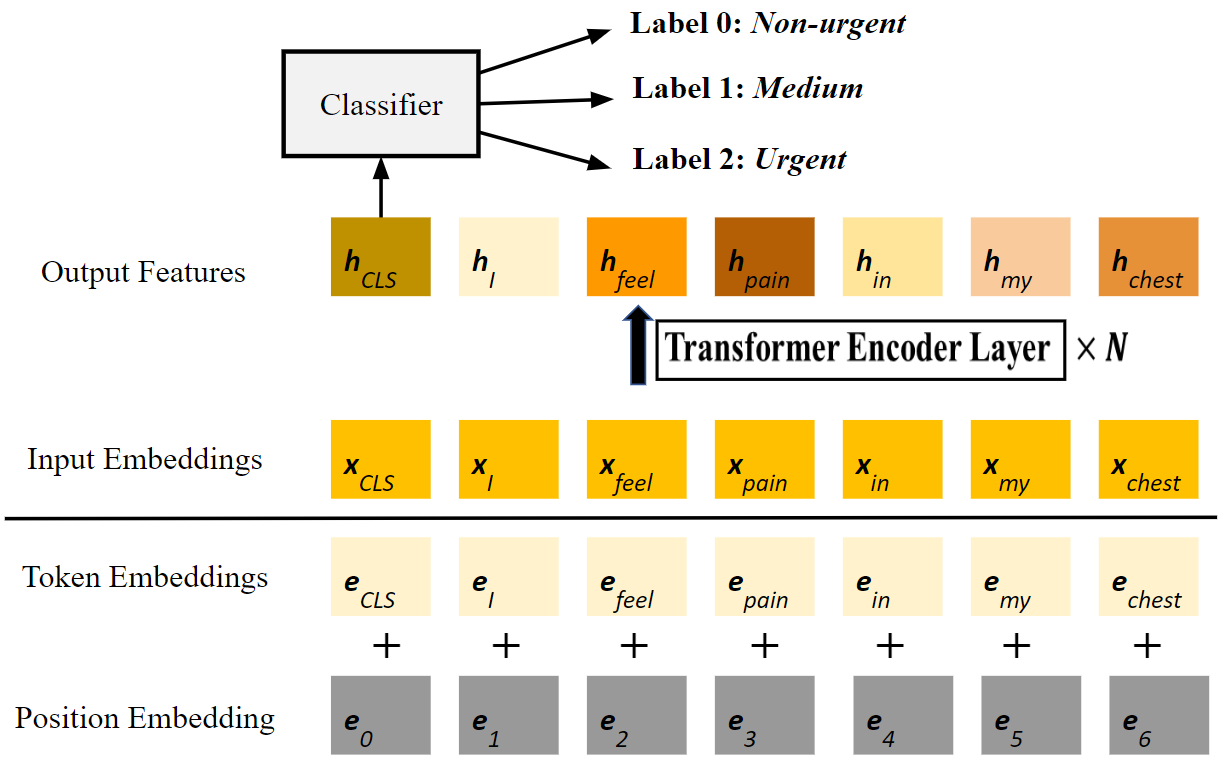}
    \end{subfigure}%
    ~
    \begin{subfigure}
        \centering
        \includegraphics[width=0.2\textwidth, height=2in]{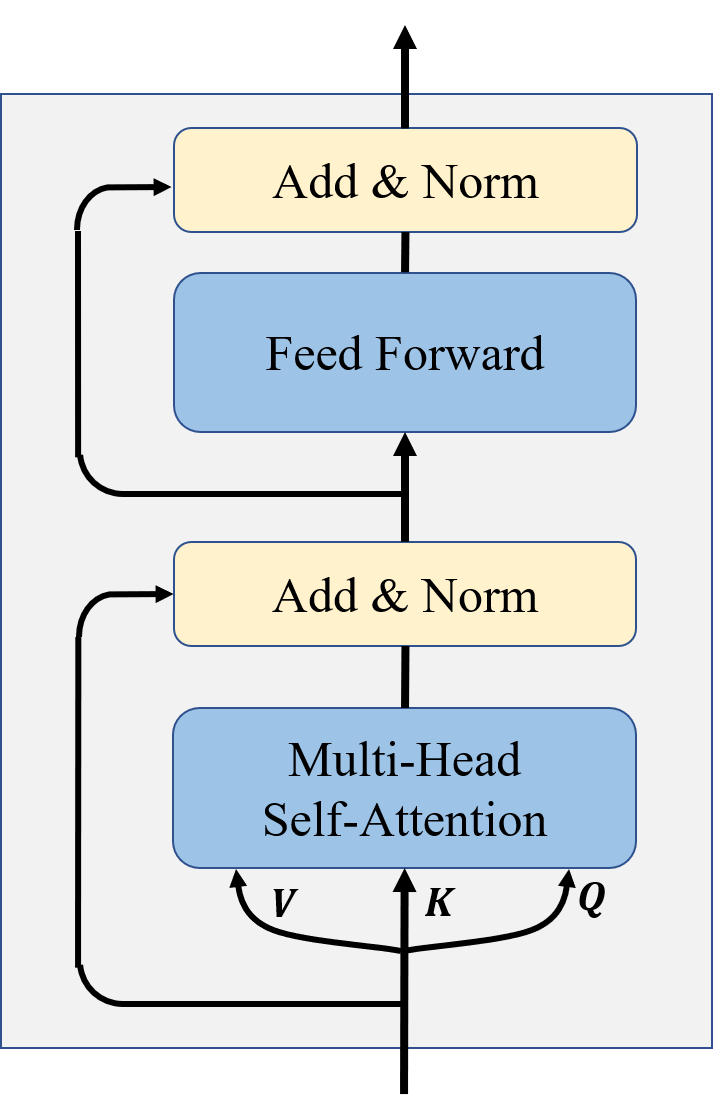}
    \end{subfigure}
    \caption{(a) BERT model structure for text classification on the message urgency dataset. Token and position embeddings are summed up as input embeddings, then fed through $N$ transformer encoder layers, yielding a high-level representations (features) for classification. (b) Architecture of one transformer encoder layer. 
    }
    \label{fig:Bert}
\end{figure*}


\section{Methods}
%
\subsection{Background}\label{sec:back}
BERT is an architecture composed of a stack of transformer encoder layers, each including two sub-layers: a \textit{multi-head self-attention} module and a \textit{feed forward} network. 
Each of these encoder layers, shown in Figure~\ref{fig:Bert}(b), is structured as a residual block with appropriate layer normalization and dropout \citep{he2016deep,ba2016layer}.
In Figure~\ref{fig:Bert}(a) we show how to fine-tune BERT-based models on the message urgency dataset to build a classifier. 
First, each text sequence of length $L$ is prepended with the special token $[CLS]$.
The sum of token and position embeddings, $[\xv_{[CLS]}, \xv_{{\rm token}_1}, \ldots, \xv_{{\rm token}_L}]$, are represented as $[\ev_{[CLS]}, \ev_{{\rm token}_1}, \ldots, \ev_{{\rm token}_L}]$ and $[\ev_0, \ldots, \ev_L]$ vectors, respectively.
We use $[\hv_{[CLS]}, \hv_{{\rm token}_1}, \ldots, \hv_{{\rm token}_L}]$ vectors to denote the output, high-level representation of classification and input tokens, each of which has the same dimensionality as the input.
After applying $N$ (usually $N=12$) different encoder layers (of the same structure in Figure~\ref{fig:Bert}(b) but different weight matrices), we obtain the high-level representations (output features) of the input sequence, in which each token representation contains information of other tokens.
Finally, the first row of the output features, $\hv_{CLS}$, is considered as the global sequence aggregator and thus fed through a softmax classification function composed of one or more fully connected layers.

The multi-head self-attention module is an ensemble of multiple attention modules sharing the same formulation. 
Given a text sequence $\tv$ represented as embedding matrix $\Xmat \in \mathbb{R}^{(L+1)\times D}$, $L$ represents the length of the text sequence and $D$ the token and position embedding dimensions.
Note that the first row in $\Xmat$ corresponds to the [$CLS$] token, hence the $L+1$ rows in $\Xmat$.
For a single-attention head, tokens of [$CLS$] and the input sequence are first mapped into the key, query and value triplets, denoted as matrices $\Kmat \in \mathbb{R}^{(L+1) \times d}$, $\Qmat\in \mathbb{R}^{(L+1) \times d}$ and $\Vmat\in \mathbb{R}^{(L+1) \times d}$, respectively, via:
\begin{equation}\label{eq:KQV}
\Kmat =  \Xmat \Wmat_K , \quad \Qmat = \Xmat \Wmat_Q , \quad \Vmat = \Xmat \Wmat_V ,
\end{equation}
where $\left\{ \Wmat_K, \Wmat_Q, \Wmat_V \right\} \in \mathbb{R}^{D \times d}$ are learnable parameters for the key, query and value of self-attention. 
With $\Kmat$, $\Qmat$ and $\Vmat$, the attention mechanism can be formulated as
\begin{align}
    \Amat & = \frac{\Qmat {\Kmat}^T}{\sqrt{d}} \in \mathbb{R}^{(L+1) \times (L+1)} \notag \\
%
    \Omat_i = {\rm attention}_i(\Kmat,&\Qmat,\Vmat)=\mbox{softmax}(\Amat) \Vmat \in \mathbb{R}^{(L+1) \times d},
    \label{eq:aggregation}
\end{align}
where $i=1,\ldots, h$, $h$ is the number of attention heads, $\mbox{softmax}(\cdot)$ is the softmax function applied row-wise.
$\Amat$ is the attention score matrix representing the compatibility of $\Kmat$ and $\Qmat$ obtained via inner products.
The multi-head self-attention is defined by concatenating and projecting the representation of each head as
\begin{equation}\label{eq:att_output}
    \Omat = [\Omat_1, \cdots, \Omat_h] \Wmat \in \mathbb{R}^{(L+1) \times D},
\end{equation}
where $[\cdot,\cdot]$ denotes column-wise concatenation and $\Wmat \in \mathbb{R}^{(d \times h) \times D}$ is a learnable projection matrix.

After the multi-head self-attention module, the position-wise (one token at the time) feed forward network module in Figure~\ref{fig:Bert}(b) composed of two fully connected layers is applied,
\begin{equation}\label{eq:ffn}
    {\rm FFN}(\xv) =  \max(0, \uv \Wmat_1 + \bv_1)  \Wmat_2 + \bv_2,
\end{equation}
where $\max(0, \cdot)$ is the standard ReLU activation function,  $\{\Wmat_1,\Wmat_2,\bv_1,\bv_2\}$ are learnable parameters, and
$\uv$ is the layer normalized residual block $\uv={\rm LayerNorm}(\xv + \ov)$, where $\xv$ (rows of $\Xmat$) and $\ov$ (rows of $\Omat$) are the inputs and outputs of the multi-head self-attention module in \eqref{eq:KQV}--\eqref{eq:att_output}, respectively, and the LayerNorm$(\cdot)$ operator is implemented according to \citep{ba2016layer}.
%

\subsection{Multi-head Attention with Label Embedding}
\begin{figure*}[t!]
    \centering
    \begin{subfigure}
        \centering
        \includegraphics[width=0.25\textwidth, height=1.5in]{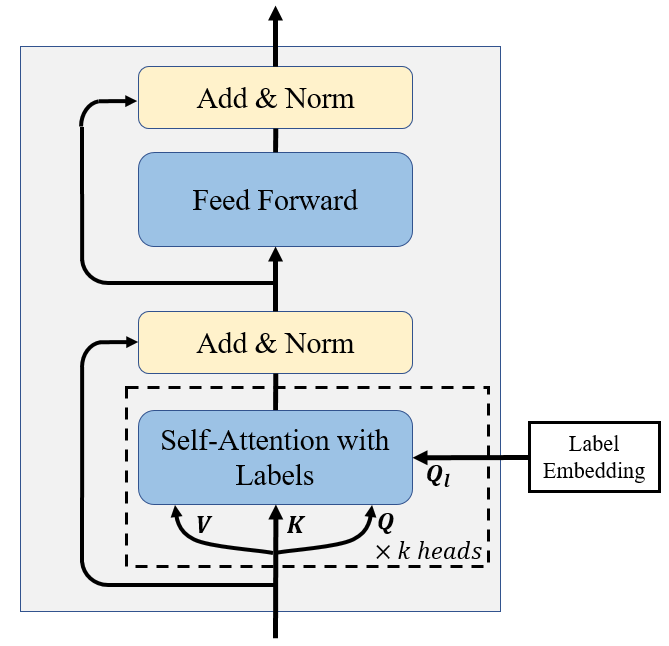}
    \end{subfigure}%
    ~ 
    \begin{subfigure}
        \centering
        \includegraphics[width=0.6\textwidth, height=1.5in]{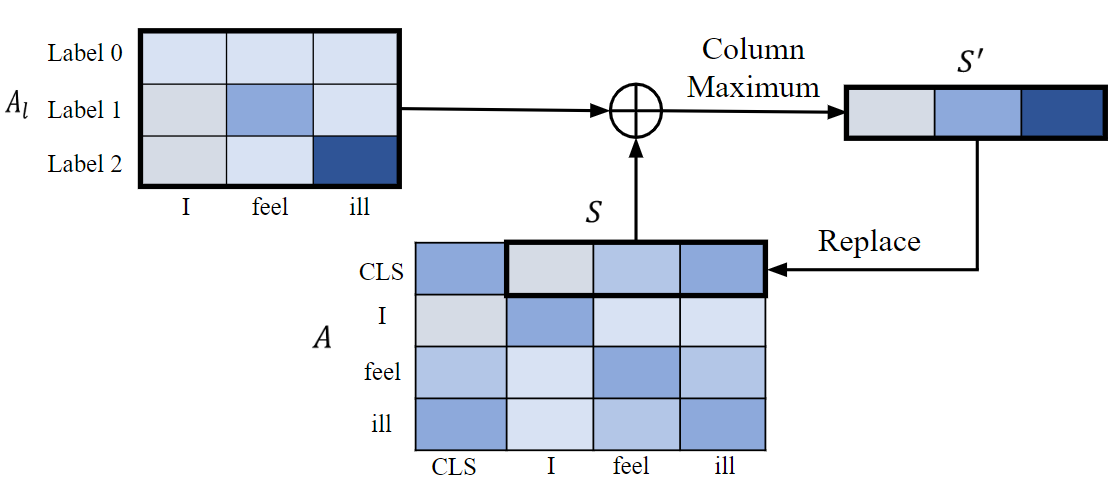}
    \end{subfigure}
    \caption{(a) Incorporating label embeddings to the multi-head self-attention in BERT. (b) Modifying self-attention scores with label embeddings. $\bigoplus$ indicates row concatenation.}
    \label{fig:m_att}
\end{figure*}

%
%
For the patient message triage task, the labeled patient text data is usually limited and hard to obtain.
Specially, data for the urgent class is scarce but of the highest priority for the  application.
Therefore, we incorporate \emph{label embeddings} as prior information into the self-attention modules, so that the model can more easily attend to class-representative keywords.
The modified module is shown in Figure~\ref{fig:m_att}(a) and described below. We asked healthcare providers to select a set of keywords or short phrases associated with the classes medium and urgent.
These are shown in Table~\ref{tab:keywords}.
The label embeddings of each class is initialized as the average of the corresponding keyword embeddings.
The label embedding for non-urgent class is initialized at randomly provided there are no immediately obvious keywords for this class.

\begin{table}[t!]
\centering
\small
\begin{tabular}{lll}
\hline\hline
Label &  Key Words/Phases \\ \hline
  Medium      &     Loss of coordination/balance, Dizziness, Near syncope, Leg swelling, Headache\\\hline
  Urgent     &    Blue lips, Chest pain, Disorientation, Paralysis, Loss of consciousness     \\\hline
\end{tabular}
\caption{Key words/phases for initialization of label embeddings.}
\label{tab:keywords}
\end{table}

Below we describe how to incorporate label embeddings into the multi-head self-attention in each
encoder layer of BERT.
Note that from Figure~\ref{fig:Bert}(a), the text sequence $\tv$ with $L$ tokens, prepended with the token $[CLS]$ as previously described, has embedding matrix $\Xmat= [\xv_{[CLS]}, \Xmat_w] \in \mathbb{R}^{(L+1) \times D}$, where the $\xv_{CLS}$ and $\Xmat_w$ are the input embedding vector for $[CLS]$ and matrix for all other input tokens in the sequence, respectively.
Subsequently, the query, key and value matrices in each attention head can be represented equivalently to~\eqref{eq:KQV} as
\begin{align}
\begin{aligned}
    &\Qmat=[\qv_{[CLS]}; \Qmat_w] = [\xv_{[CLS]}; \Xmat_w]\Wmat_Q\in \mathbb{R}^{(L+1) \times d},\\
    &\Kmat=[\kv_{[CLS]}; \Kmat_w] = [\xv_{[CLS]}; \Xmat_w]\Wmat_K \in \mathbb{R}^{(L+1) \times d}, \\
    &\Vmat=[\vv_{[CLS]}; \Vmat_w] = [\xv_{[CLS]}; \Xmat_w]\Wmat_V \in \mathbb{R}^{(L+1) \times d},
    \label{eq:encode}
\end{aligned}
\end{align}
where $[\cdot;\cdot]$ denotes row-wise concatenation, and $\{\qv_{[CLS]}, \kv_{[CLS]}, \vv_{[CLS]}\}$ and $\{\Qmat_w, \Kmat_w, \Vmat_w\}$ are the query, key and value triplets of the $[CLS]$ token and the other input tokens, respectively. 
From \eqref{eq:aggregation}, the original attention weights (before softmax) of the augmented sequence can be written as
\begin{equation}\label{eq:bert_att}
\Amat_{\rm BERT} = \frac{1}{\sqrt{d}}
\begin{bmatrix}
\qv_{[CLS]}^T \\
\Qmat_w
\end{bmatrix}
\begin{bmatrix}
\kv_{[CLS]}& \Kmat_w^T 
\end{bmatrix}
=\frac{1}{\sqrt{d}}
\begin{bmatrix}
\qv^T_{[CLS]}\kv_{[CLS]}& \qv^T_{[CLS]}\Kmat_w^T \\
\Qmat_W\kv_{[CLS]} & \Qmat_W\Kmat_w^T
\end{bmatrix}.
\end{equation}
In $\Amat_{\rm BERT}$, which is equivalent to \eqref{eq:aggregation}, the cross-attention
between the $[CLS]$ token and all input tokens is denoted as $\Smat\triangleq\qv^T_{[CLS]}\Kmat_w^T \in \mathbb{R}^{1 \times L}$.

In LESA-BERT, we introduce label embeddings to boost the model's attention to keywords associated with different labels. We incorporate label embeddings to self-attention in three steps.
First (Step 1), we compute the cross attention between the label embeddings and the message tokens
\begin{align}
    \vspace{-0.3in}
    \Qmat_l &= \Xmat_l\Wmat_Q \in \mathbb{R}^{3 \times d}, \\
    \Amat_l &= \frac{\Qmat_l\Kmat^T_w}{\sqrt{d}} \in \mathbb{R}^{3 \times L},
\end{align}
where $\Xmat_l\in \mathbb{R}^{3 \times D}$ is a matrix containing the three label embeddings, \emph{i.e.}, non-urgent, medium and urgent, which are encoded into label queries $\Qmat_l$ via the same $\Wmat_Q$ as in~\eqref{eq:encode}.
Next (Step 2), we compute a modified cross-attention row vector $\Smat^\prime$ as
\begin{equation}
    \Smat^\prime = \max([\Smat; \Amat_l]) \in \mathbb{R}^{1 \times L},
    \label{eq:s_prime}
\end{equation}
where we concatenate $\Smat$ and $\Amat_l$ by row
and then keep the maximum value of each column.
As a result, $\Smat^\prime$ represents the maximum attention score of a input token with both the $[CLS]$ token and the label embeddings.
Finally (Step 3), we obtain the attention weights in LESA-BERT by replacing $\Smat$ by $\Smat^\prime$ in~\eqref{eq:bert_att}, thus obtaining
\begin{equation}\label{eq:lesa_att}
\Amat_{\rm LESA-BERT} = \frac{1}{\sqrt{d}}
\begin{bmatrix}
\qv^T_{[CLS]}\kv_{[CLS]}& \Smat^\prime \\
\Qmat_W\kv_{[CLS]} & \Qmat_W\Kmat_w^T
\end{bmatrix}.
\end{equation}
%
In~\eqref{eq:lesa_att}, when a token is highly relevant to one of the labels it will result in a larger attention score with $[CLS]$ in $\Smat^\prime$, thus the $[CLS]$ embedding will be less affected by irrelevant information in the sequence, unlike \eqref{eq:aggregation} where only attention from the current $[CLS]$ embedding is considered.
The proposed attention layer is shown in Figure \ref{fig:m_att}(b).
The attention score matrix $\Amat$ in~\eqref{eq:aggregation} is replaced as $\Amat_{\rm LESA-BERT}$ in ~\eqref{eq:lesa_att}.
All other components are exactly the same as the original encoders in BERT as shown in~\eqref{eq:KQV}--\eqref{eq:ffn}.

Note that by incorporating label embeddings to the self-attention layer in
BERT model, \eqref{eq:s_prime} allows the model to account for label information across all layers.
We share the same label embedding for all the layers. The label embedding is adapted to different layers via each layer's $\Wmat_Q$s in the multi-head attention module. 
The label embeddings include trainable parameters, which are tuned simultaneously with other parameters in BERT.
If keywords for labels are available, then label embeddings are initialized accordingly, otherwise, label embeddings could also be initialized at random \citep{wang2018joint}. 
All other parameters can be initialized from the pre-trained Bert or BioBert model.
Though LESA-BERT is motivated by our application, it can be employed for general text classification tasks.

To the best of the authors' knowledge, LESA-BERT is the first work to incorporate label embeddings to perform self-attention in BERT encoders.
Note that this technique could also be applied to models that have the self-attention components such as GPT-2, and transformers \citep{radford2019language}.

\subsection{Knowledge Transfer with Model Distillation}
\begin{figure}[t!]
\centering
\includegraphics[width=8cm]{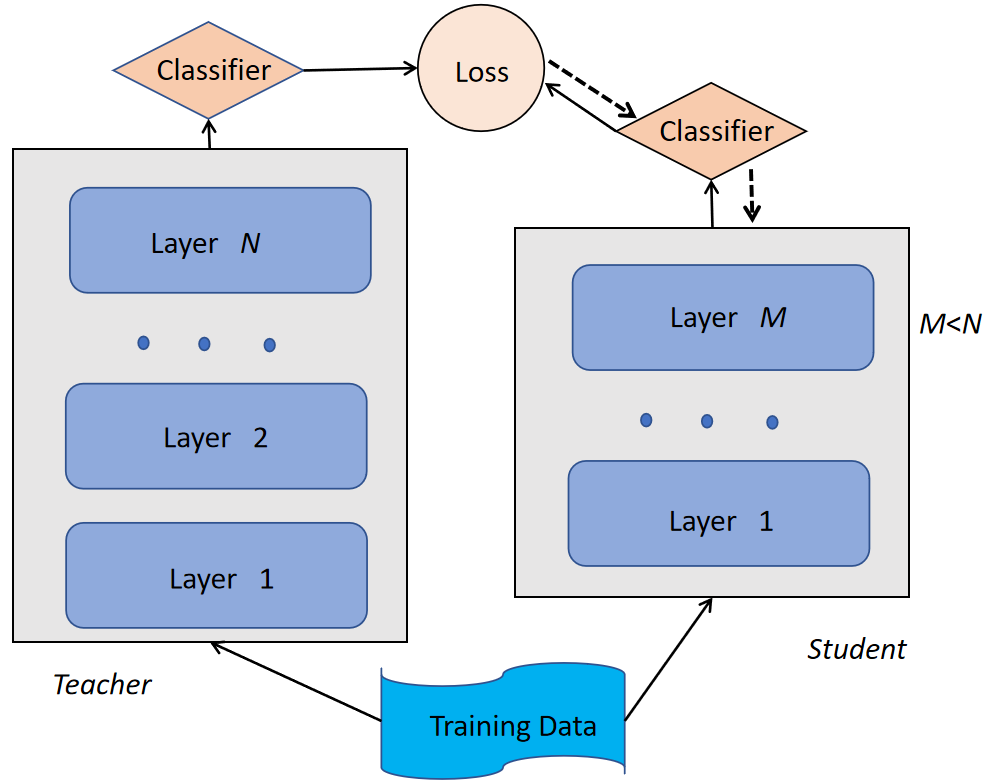}
\caption{Example of knowledge distillation from a N-layered (teacher) to a M-layered (student) model with $M<N$. In our case, each layer is a
LESA-BERT encoder. Solid black lines represent forward computations, whereas dashed lines indicate error back-propagation. The loss function is $L_{ce}$ in~\eqref{eq:distil_loss}. 
\label{fig:distilBert}}
\end{figure}

Knowledge distillation \citep{bucilua2006model,hinton2015distilling} is a 
compression technique, in which a compact model, called the student model, is trained to reproduce the behavior of a larger teacher model, as in Figure \ref{fig:distilBert}.
In our experiments, we use a loss function defined over the cross-entropy between the teacher and the student class probabilities, given  by:
\begin{equation}\label{eq:distil_loss}
    L_{ce} = -\text{softmax}(\bm{z}^t/T_0)^T\cdot \log \text{softmax}(\bm{z}^s/T_0),
\end{equation}
where $\bm{z}^t$ and $\bm{z}^s$ are vectors of the output logits from the teacher and student classification networks, respectively, and $T_0$ is a parameter controlling 
the degree to which we focus on the class with the highest probability, usually set to 1.
When the probabilities ${\rm softmax}(\bm{z}^t/T_0)$ from the teacher model are similar to those of the student model, the cross entropy loss in \eqref{eq:distil_loss} is small.
Therefore, the cross entropy loss measures the difference between output
probabilities of teacher and student classifiers.

The knowledge distillation technique is typically used to compress large models such as BERT in order to reduce their computational and memory costs; usually to allow their deployment \citep{bucilua2006model,hinton2015distilling, sanh2019distilbert,tang2019distilling, strubell2019energy}.
Here, we hypothesize that, beyond these advantages, knowledge
distillation can 
reduce overfitting in the case of small datasets as is our case.
Specifically, we expect a smaller generalization error of the distilled model due to its smaller size. According to this hypothesis, we propose a distilled variant of the LESA-BERT: a small BERT model with fewer encoder layers (student) trained to reproduce the behavior of the fine-tuned LESA-BERT (teacher), which has 12 layers. 
In our experiments, we indeed show the improved performance of the distilled model.

\section{Experimental Results}
%

\subsection{Cohort}\label{sec:data}
%

%
In this work, we utilized 1,756 web portal messages generated from 10/2014 to 08/2018 by adult patients ($>18$ years old) of a large academic medical center. 
The Electronic Health Record (EHR) system (Epic Verona, WI, USA) with associated patient portal (MyChart) was the source of all patient messages.
A custom-built Application Programming Interface (API) securely made available the portal messages from the EHR enterprise data warehouse into a highly protected virtual network space offered by the medical center.
Approved users were allowed access to work with the identifiable protected health information. 
These messages included free, unstructured plain text sent by patients to their healthcare team. 
Responses and messages sent from the clinician or health system to the patient were excluded from the analysis. 
Portal messages were manually labeled by experienced sub-specialty (cardiology) clinicians into three levels of priority: non-urgent, medium and urgent.
Non-urgent labels include notes of appreciation (\emph{e.g.}, thank you).
The Medium urgency class contains messages that could be reasonably responded to in 1-3 days.
Urgent messages are those requiring an immediate phone call to the patient by the clinician. 
Conditions suggesting acute myocardial infarction, exacerbation of heart failure respiratory distress or possible stroke were labeled as urgent and would be inappropriate for an asynchronous 
patient portal.

As summarized in Table \ref{tab:data}, the data set is imbalanced.
Specifically, the total number of messages, 1,756, includes 631 non-urgent, 955 medium and 170 urgent messages.
Urgent messages are scarce ($\sim10\%$), thus the culprit of the imbalance issue.
Table \ref{tab:data} also includes a typical example message for each class.
For example, an urgent message could be that a patient reports chest
pain. 
In our experiments, the dataset is split into 80\% training set ($\sim1.4$K)
and 20\% test set ($\sim0.35$K). 







\subsection{Baselines}\label{sec:baseline}
To evaluate the performance of LESA-BERT, we compare it to strong baseline classifiers of three kinds: ($i$) traditional machine learning methods: SVM, ($ii$) shallow neural networks: text CNN and Bi-LSTM with attention layer, and  ($iii$) pre-trained deep learning models: BERT and BioBERT.



\textbf{SVM:} We utilize an $\ell_2$ regularized SVM classifier \citep{evgeniou2000regularization}, which optimizes the hinge loss function with a $\ell_2$ penalty.
SVMs are very effective for high-dimensional data \citep{ghaddar2018high}, especially when the number of dimensions is greater  than that of samples.
SVMs have been widely used for text classification \citep{tong2001support,dadgar2016novel}.
In our experiments, we implement linear SVM classifiers with the Python module scikit-learn \citep{scikit-learn}, and train it via stochastic gradient descent (SGD) with $\ell_2=6\times10^{-4}$ penalty coefficient.
The learning rate is set by \citet{bottou2010large} and the weights
are randomly initialized.

\textbf{Bi-LSTM with Attention:} LSTM, short for long short term memory, is one kind of neural network suited for sequential data.
In NLP, bidirectional (forward and backward) LSTMs are usually used as feature extractors for sequences of tokens.
On top of the Bi-LSTM layer, the attention layer is introduced to capture important words that drive the decisions of the document classification \citep{yang2016hierarchical}.
The attention weights are further employed to compute the weighted sum of output 
vectors of Bi-LSTM as the hidden representation for each message.
Then, this representation is fed into a fully connected layer that produces the logits for the three labels.
The hidden dimension of Bi-LSTM is set to 60, and the max length of each message is set to 256 tokens.
We implement Bi-LSTM with attention model in Pytorch \citep{paszke2019pytorch}, and train it using Adam \citep{kingma14adam} with batch size of 8. The learning rate is $0.01$ and the weight
matrices are initialized with method in \citet{he2015delving}.

\textbf{Text CNN:} Details of text CNN model can be found in \citet{kim2014convolutional}. We set the following parameters for our CNN model: the convolution kernel sizes: $\{1, 2, 3, 4, 5\}$, the filter numbers: $\{200, 300, 500, 500, 200\}$,
dropout probability: 0.5, and maximum number of tokens in each message
is 256. 
Text CNN also comprises maximum pooling layers, as well as the rectified linear unit (RELU) used as the non-linear activation function. The parameters of the network are learned in mini-batches of size 8 using Adam \citep{kingma14adam}. The learning rate is $0.01$ and the weight
matrices are initialized with method in \citet{he2015delving}.

\textbf{BERT/BioBERT:} For BERT and BioBERT, the base uncased, \emph{i.e.}, all words are treated as lowercase, model is employed in this research. 
BERT base comprises $12$ layers, $768$ hidden units, $12$ self-attention heads, and $110$M parameters. These model are trained with warm-up Adam \citep{gotmare2018a} setting a batch size of 8 and a learning rate of
$3\times 10^{-5}$. The maximum sequence length is set at 256 and other
parameters remain the same as the default BERT base configuration.
The weight matrices in BERT and BioBERT are set to BERT base uncased
\citep{devlin2018bert}
and BioBERT v1.1 \citep{lee2020biobert}, respectively.

\subsection{Proposed Models}
We briefly cover the configuration of our models, which 
are implemented with Pytorch in a Python 3 environment\footnote{Source code for our models will be publicly released online upon acceptance.}.

\textbf{LESA-BERT:} LESA-BERT has the same configuration of BioBERT except for its label embeddings, where embeddings of urgent and medium labels are initialized by the average embeddings of their keywords (see Table~\ref{tab:keywords}) while the embeddings of non-urgent labels is randomly initialized.

\textbf{Distil-LESA-BERT:} We distill the fine-tuned LESA-BERT model
to two variants: one with 6 encoder layers (Distil-LESA-BERT-6), and the another one with 3 (Distil-LESA-BERT-3).
The learning rate for these two variants is $2\times10^{-5}$ and they are initialized from the first 6 or 3 layers of the pre-trained BioBert model accordingly.

\subsection{Feature Choices}
We follow the literature to choose features of each message for the classification task \citep{sulieman2017classifying}.
For traditional classifiers like SVM, TF-IDF features are commonly used.
For shallow neural networks like TextCNN and Bi-LSTM with attention, word embeddings
such as Word2vec or GloVe are utilized as input features.
For BERT-based deep learning models, we just tokenize each message then the model initializes them from pre-training embeddings \citep{devlin2018bert}.


For the SVM classifier, we use TF-IDF vectors \citep{salton1988term} of each message as features.
TF represents the frequency of a certain token in a message and IDF represents the inverse of the number of messages in which this token has appeared.
The product of TF and IDF for a token is a score that represents the importance of the token in a message.
TF-IDF features have been shown to be very effective in text classification \citep{zhang2008tfidf}.

For shallow neural network classifiers like CNN or LSTM models, instead of using TF-IDF, we adopt pre-trained GloVe word vectors \citep{pennington2014glove} as features for each message. 
We use the 100 dimensional GloVe word vectors pre-trained on 6 billion tokens from Wikipedia and Gigaword data sets.

For the BERT-based models including BERT, BioBERT and LESA-BERT, we use the wordpiece tokenizer \citep{wu2016google} and keep the models' pre-trained token
embeddings as initialization, so that the input embeddings are compatible with the
models' pre-trained parameters. We utilize the BERT base uncased model with 12 layer encoders
and 12 attention heads and 768 dimensional embeddings, which was 
pre-trained on BookCorpus \citep{zhu2015aligning} and English Wikipedia.
BioBERT shares the same model configuration as BERT base, but was further pre-trained
on biomedical corpora (PubMed abstracts and PMC full-text articles).
Parameters in LESA-BERT except the label embeddings are initialized from the pretrained BioBert model.

\subsection{Evaluation Metrics} 


For classification tasks, commonly utilized metric criteria are precision, recall, F1 score and Area Under the Curve (AUC) score.
Oftentimes a classifier has a good precision with a poor recall, thus F1 score provides a
good balance of precision and recall.
AUC is typically employed for binary classifiers.
Since the message urgency dataset comprises multiple (three) classes and is imbalanced, we find that macro average F1 score as the most suitable evaluation criteria \citep{parambath2014optimizing}.
The F1 score is given by the harmonic mean of precision and recall.
Accuracy is a poor metric in this situation because it encourages the model to focus on the majority class.
Similarly, micro-level metrics like micro F1 score are not preferred because they are weighted by the number of cases in each label, which makes them favor the majority classes.

We also evaluate the advantages of the distillation process.
We distill the fine-tuned LESA-BERT model to 6-layered and 3-layered versions, and evaluate the number of parameters and the inference time required for a full pass on the test set on a CPU with batch size of 1, which are commonly used evaluation criteria for distillation \citep{sanh2019distilbert}. 



\begin{table}[t!]
  \centering
      \caption{Performance metrics of different classifiers on the patient messages data.}
  \begin{tabular}{lcccc}\hline\hline
    Model & Macro F1 & Macro Precision & Macro Recall \\ \hline
    SVM & $0.748\pm0.007$ & $0.795\pm0.007$ & $0.731\pm0.006$  \\ \hline 
    TextCNN & $0.754\pm0.020$ & $0.772\pm0.031$ & $0.749\pm0.031$  \\ 
    Bi-LSTM Attention & $0.761\pm0.016$ & $0.758\pm0.016$ & $0.769\pm0.021$  \\ \hline 
    BERT & $0.761\pm0.021$ & $0.762\pm0.019$ & $0.761\pm0.024$ \\      
    BioBERT & $0.764\pm0.010$ & $0.774\pm0.015$ & $0.758\pm0.009$  \\ 
    \hline
    LESA-BERT & $\bm{0.789}\pm\bm{0.011}$ & $0.784\pm0.010$ & $0.797\pm0.014$  \\ 
    Distil-LESA-BERT-6 & $\bm{0.807}\pm\bm{0.009}$ & $0.816\pm0.004$ & $0.798\pm0.024$  \\ 
    Distil-LESA-BERT-3 & $\bm{0.780}\pm\bm{0.017}$ & $0.768\pm0.016$ & $0.816\pm0.015$ \\ \hline
  \end{tabular}
  \label{tab:result} 
\end{table}

\subsection{Quantitative Results} 
%
Table~\ref{tab:result} presents quantitative results for various classifiers on the test set data.
Mean and standard errors are computed based on 5 different random seeds for the SGD/Adam training algorithm.
The F1 scores for shallow neural networks, \emph{i.e.}, TextCNN and Bi-LSTM attention,
are slightly better than SVM, which might be owed to the benefits of semantic features captured by GloVe word vectors.
BERT-based deep neural networks such as BERT and BioBERT only marginally outperforms Bi-LSTM attention and TextCNN, which is likely caused by the small size ($\sim 1.4$K) of the training data.
The F1 score of BioBERT is 0.3\% higher than that of BERT, which means that pre-training on large biomedical corpora transfers extra beneficial information to BERT model.
In terms of macro average F1 score, our LESA-BERT and its distilled variants achieve higher F1 scores compared with other classifiers.
The performance gains can be attributed to the addition of the label embeddings.

The model Distil-LESA-BERT-6 provides higher F1 score (0.807) than that of the full 12-layered LESA-BERT.
Possibly, this implies that the full model overfits the training set, while the distilled model with 6-layers reduces the overfitting.
Distil-LESA-BERT-3 model, which comprises only 3 layers provides F1 score of 0.78, slightly lower than the full LESA-BERT model.


\subsection{Computational Cost}
Table~\ref{tab:time_params} presents the number of parameters in LESA-BERT and its distilled variants, as well as inference times for a full pass over the test set with batch size of 1.
As the number of layers in LESA-BERT decreases, the number of parameters and inference time reduces dramatically.
Distilling the original 12-layered LESA-BERT to 6 layers reduces the number of parameters by 40\% and accelerates the inference by 2.7 times. 
Distil-LESA-BERT-3 has fewer parameters and faster inference speed than Distil-LESA-BERT-6.
Taking into account their F1 scores, Distil-LESA-BERT-6 outperforms the LESA-BERT, and Distil-LESA-BERT-3 retains 98.9\% of LESA-BERT performance.
Distil-LESA-BERT-6 provides the best F1 score compared to all other methods.
However, in the case that inference time is preferred, Distil-LESA-BERT-3 will be the top choice. 

\begin{table}[t!]
\centering
\caption{Number of parameters and inference times on the patient messages data.}
\begin{tabular}{lll}
\hline\hline
Moel         & \# param.  & Inf. time \\
             & (Millions) & (Seconds) \\ \cline{2-3} 
LESA-BERT   & $110(\times 1.0)$        & $212.6(\times 1.0)$     \\
Distil-LESA-BERT-6 & $66(\times 0.6)$         & $79.8(\times 2.7)$      \\
Distil-LESA-BERT-3 & $44(\times 0.4)$         & $40.8(\times 5.2)$      \\ \hline
\end{tabular}
\label{tab:time_params}
\end{table}

\subsection{Qualitative Results}
\begin{figure}[t!]
    \centering
    \begin{subfigure}
        \centering
        \includegraphics[width=0.95\textwidth]{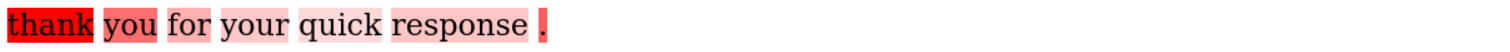}
    \end{subfigure}%
    \begin{subfigure}
        \centering
        \includegraphics[width=0.95\textwidth]{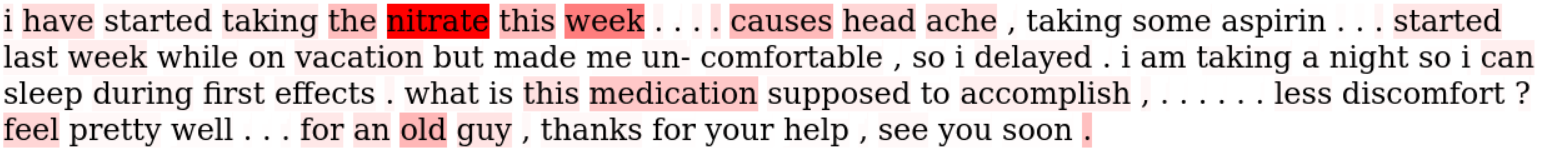}
    \end{subfigure}
    \begin{subfigure}
        \centering
        \includegraphics[width=0.95\textwidth]{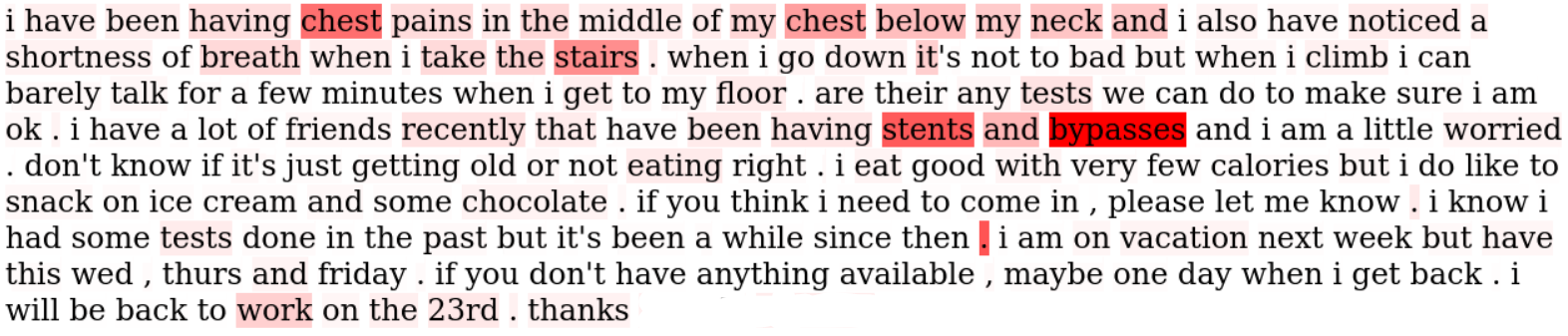}
    \end{subfigure}
    \caption{Visualization of learned LESA-BERT attention scores for tokens in messages in test data. One example from each label: (a) non-urgent, (b) medium, and (c) urgent.}
    \label{fig:label_attention}
\end{figure}

In Figure~\ref{fig:label_attention}, we visualize how the obtained representation by LESA-BERT attend to different tokens in three example messages, one for each class.
Tokens are shaded in terms of their importance for classification.
Dark-colored tokens are more prominently weighted (attended) when constructing the message embedding.
As seen in Figure~\ref{fig:label_attention}, attention scores can identify relevant keywords in all three examples.
We choose these three examples because they are representative in their respective categories.



\begin{figure}[h!]
\centering
\includegraphics[width=8cm]{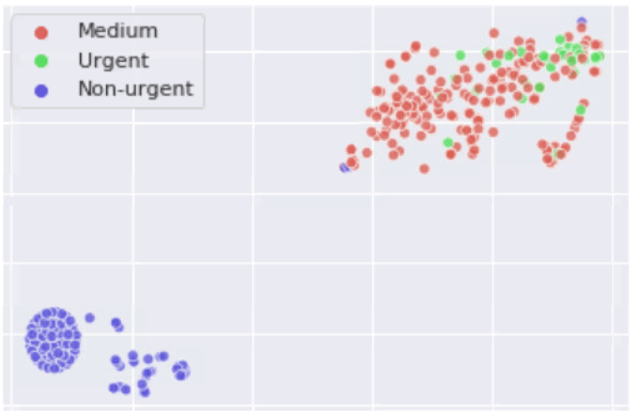}
\caption{$t$-SNE representation of [$CLS$] tokens for messages grouped by class. 
\label{fig:cluster}}
\end{figure}

Figure~\ref{fig:cluster} shows the 2-dimensional $t$-SNE plot \citep{maaten2008visualizing} of the latent code of each message in the test set, \emph{i.e.}, the embedding of [$CLS$] token in the output encoder of LESA-BERT.
Each point corresponds to a single message and the colors represent the different labels.
From this figure, non-urgent messages (purple dots) are well separated from messages of the other two classes.
For the urgent and medium classes, their data points have significant overlap, meaning
that urgent and medium messages, not surprising, have similar characteristics (features).
This may be in part caused by the subjectiveness of the labeling process as the definitions of urgent and medium messages are not exactly clear.

%

\section{Discussion}
We have proposed a framework for building classifiers on small and imbalanced datasets, which is a commons scenario in healthcare.
Our method builds upon a large deep learning model, BioBERT, which was pre-trained on huge general-domain and biomedical corpora.
We developed LESA-BERT, which has a novel self-attention architecture that can incorporate label embeddings to boost the model's capacity for attention.
We fine-tuned LESA-BERT on the target dataset by initializing its parameters from BioBERT.
Subsequently, we distilled the fine-tuned 12-layered LESA-BERT to 6-layered or 3-layered variants, and found that the 6-layered Distil-LESA-BERT outperforms the 12-layered LESA-BERT.
Therefore, knowledge distillation is not only a tool for model compression, but can also be used to reduce overfitting. 
We demonstrated the application of our framework on a real healthcare dataset --\emph{message urgency}-- and built a message triage classification model.
Our methods outperformed baseline classifiers by a significant margin.
Our technical solution can be easily applied to other clinical datasets.

\paragraph{Clinical Implications}
We built an automatic message triage model that can predict the priority of patient portal messages based on their content.
This version of message triage could be further improved with a larger dataset to train our classifiers.
The message triage system is of real impact to healthcare providers and potentially save valuable working hours by freeing them from reading patient portal messages in chronological order.
Promptly accessing and realizing urgent messages can impact both patient safety as well as clinician workflow.
After the message triage classifier produces the priority of messages, healthcare providers can take further actions based on their existing workflow.
Further, they may want to utilize automatic responses or templates to answer non-urgent messages.
Medium or time-sensitive messages may either be directed to emergency rooms, urgent care or telephone encounters.
%

\paragraph{Limitations}
%
Although our methods outperform a wide selection of baseline models, it can still produce some mistakes from a human perspective.
For instance, Figure~\ref{fig:cluster} shows that two non-urgent messages (purple dots) fall in the region of medium and urgent labels.
In fact, we have investigated these very few messages that are close to medium or urgent messages, and one example reads ``No chest pain at all".
This message has very strong keywords, chest pain, which usually appears in urgent messages.
Therefore, the LESA-BERT model maps this message to urgent class by ignoring the negation word ``No". Many researchers have found that deep learning models are powerful but still easy to fool
\citep{moosavi2016deepfool,heaven2019deep}.
As future work we plan on incorporating negation detection \citep{qian2016speculation} into our classification framework.

Our current message triage system has a few limitations from the clinical perspective.
Firstly, our message classifier only has three levels of urgency, which could be further refined into more granular categories. 
After passing patient portal messages through our urgency classifier, urgent messages will be treated with priority, but non-urgent and medium ones will probably need further automatic classifiers, such as message intention detector, \emph{etc}.
Secondly, our current work is limited to only utilizing the content of portal messages.
One possible follow-up research could be to incorporate patient's demographic information or even comorbidities.
For example, age can be quite important when deciding the priority of messages.
Messages from patients aged over 65 should receive relatively more immediate attention than younger patients.
Comorbidities could also be taken into account as patients with heart attacks, for example, usually get more intense care than patients with, for example, common respiratory infections.


\bibliography{reference}

\begin{thebibliography}{52}
\providecommand{\natexlab}[1]{#1}
\providecommand{\url}[1]{\texttt{#1}}
\expandafter\ifx\csname urlstyle\endcsname\relax
  \providecommand{\doi}[1]{doi: #1}\else
  \providecommand{\doi}{doi: \begingroup \urlstyle{rm}\Url}\fi

\bibitem[Akata et~al.(2015)Akata, Perronnin, Harchaoui, and
  Schmid]{akata2015label}
Zeynep Akata, Florent Perronnin, Zaid Harchaoui, and Cordelia Schmid.
\newblock Label-embedding for image classification.
\newblock \emph{IEEE transactions on pattern analysis and machine
  intelligence}, 38\penalty0 (7):\penalty0 1425--1438, 2015.

\bibitem[Ba et~al.(2016)Ba, Kiros, and Hinton]{ba2016layer}
Jimmy~Lei Ba, Jamie~Ryan Kiros, and Geoffrey~E Hinton.
\newblock Layer normalization.
\newblock \emph{arXiv preprint arXiv:1607.06450}, 2016.

\bibitem[Bottou(2010)]{bottou2010large}
L{\'e}on Bottou.
\newblock Large-scale machine learning with stochastic gradient descent.
\newblock In \emph{Proceedings of COMPSTAT'2010}, pages 177--186. Springer,
  2010.

\bibitem[Bucilu{\v a} et~al.(2006)Bucilu{\v a}, Caruana, and
  Niculescu-Mizil]{bucilua2006model}
Cristian Bucilu{\v a}, Rich Caruana, and Alexandru Niculescu-Mizil.
\newblock Model compression.
\newblock In \emph{Proceedings of the 12th ACM SIGKDD international conference
  on Knowledge discovery and data mining}, pages 535--541, 2006.

\bibitem[Chen et~al.(2019)Chen, Lalor, Liu, Druhl, Granillo, Vimalananda, and
  Yu]{chen2019detecting}
Jinying Chen, John Lalor, Weisong Liu, Emily Druhl, Edgard Granillo, Varsha~G
  Vimalananda, and Hong Yu.
\newblock Detecting hypoglycemia incidents reported in patients’ secure
  messages: Using cost-sensitive learning and oversampling to reduce data
  imbalance.
\newblock \emph{Journal of medical Internet research}, 21\penalty0
  (3):\penalty0 e11990, 2019.

\bibitem[Cronin et~al.(2015)Cronin, Fabbri, Denny, and
  Jackson]{cronin2015automated}
Robert~M Cronin, Daniel Fabbri, Joshua~C Denny, and Gretchen~Purcell Jackson.
\newblock Automated classification of consumer health information needs in
  patient portal messages.
\newblock In \emph{AMIA Annual Symposium Proceedings}, volume 2015, page 1861.
  American Medical Informatics Association, 2015.

\bibitem[Cronin et~al.(2017)Cronin, Fabbri, Denny, Rosenbloom, and
  Jackson]{cronin2017comparison}
Robert~M Cronin, Daniel Fabbri, Joshua~C Denny, S~Trent Rosenbloom, and
  Gretchen~Purcell Jackson.
\newblock A comparison of rule-based and machine learning approaches for
  classifying patient portal messages.
\newblock \emph{International journal of medical informatics}, 105:\penalty0
  110--120, 2017.

\bibitem[Dadgar et~al.(2016)Dadgar, Araghi, and Farahani]{dadgar2016novel}
Seyyed Mohammad~Hossein Dadgar, Mohammad~Shirzad Araghi, and Morteza~Mastery
  Farahani.
\newblock A novel text mining approach based on tf-idf and support vector
  machine for news classification.
\newblock In \emph{2016 IEEE International Conference on Engineering and
  Technology (ICETECH)}, pages 112--116. IEEE, 2016.

\bibitem[Devlin et~al.(2018)Devlin, Chang, Lee, and Toutanova]{devlin2018bert}
Jacob Devlin, Ming-Wei Chang, Kenton Lee, and Kristina Toutanova.
\newblock Bert: Pre-training of deep bidirectional transformers for language
  understanding.
\newblock \emph{arXiv preprint arXiv:1810.04805}, 2018.

\bibitem[Evgeniou et~al.(2000)Evgeniou, Pontil, and
  Poggio]{evgeniou2000regularization}
Theodoros Evgeniou, Massimiliano Pontil, and Tomaso Poggio.
\newblock Regularization networks and support vector machines.
\newblock \emph{Advances in computational mathematics}, 13\penalty0
  (1):\penalty0 1, 2000.

\bibitem[Ghaddar and Naoum-Sawaya(2018)]{ghaddar2018high}
Bissan Ghaddar and Joe Naoum-Sawaya.
\newblock High dimensional data classification and feature selection using
  support vector machines.
\newblock \emph{European Journal of Operational Research}, 265\penalty0
  (3):\penalty0 993--1004, 2018.

\bibitem[Goldzweig et~al.(2013)Goldzweig, Orshansky, Paige, Towfigh, Haggstrom,
  Miake-Lye, Beroes, and Shekelle]{goldzweig2013electronic}
Caroline~Lubick Goldzweig, Greg Orshansky, Neil~M Paige, Ali~Alexander Towfigh,
  David~A Haggstrom, Isomi Miake-Lye, Jessica~M Beroes, and Paul~G Shekelle.
\newblock Electronic patient portals: evidence on health outcomes,
  satisfaction, efficiency, and attitudes: a systematic review.
\newblock \emph{Annals of internal medicine}, 159\penalty0 (10):\penalty0
  677--687, 2013.

\bibitem[Gotmare et~al.(2019)Gotmare, Keskar, Xiong, and Socher]{gotmare2018a}
Akhilesh Gotmare, Nitish~Shirish Keskar, Caiming Xiong, and Richard Socher.
\newblock A closer look at deep learning heuristics: Learning rate restarts,
  warmup and distillation.
\newblock In \emph{International Conference on Learning Representations}, 2019.
\newblock URL \url{https://openreview.net/forum?id=r14EOsCqKX}.

\bibitem[He et~al.(2015)He, Zhang, Ren, and Sun]{he2015delving}
Kaiming He, Xiangyu Zhang, Shaoqing Ren, and Jian Sun.
\newblock Delving deep into rectifiers: Surpassing human-level performance on
  imagenet classification.
\newblock In \emph{Proceedings of the IEEE international conference on computer
  vision}, pages 1026--1034, 2015.

\bibitem[He et~al.(2016)He, Zhang, Ren, and Sun]{he2016deep}
Kaiming He, Xiangyu Zhang, Shaoqing Ren, and Jian Sun.
\newblock Deep residual learning for image recognition.
\newblock In \emph{Proceedings of the IEEE conference on computer vision and
  pattern recognition}, pages 770--778, 2016.

\bibitem[Heaven(2019)]{heaven2019deep}
Douglas Heaven.
\newblock Why deep-learning ais are so easy to fool.
\newblock \emph{Nature}, 574\penalty0 (7777):\penalty0 163, 2019.

\bibitem[Hefner et~al.(2019)Hefner, MacEwan, Biltz, and
  Sieck]{hefner2019patient}
Jennifer~L Hefner, Sarah~R MacEwan, Alison Biltz, and Cynthia~J Sieck.
\newblock Patient portal messaging for care coordination: a qualitative study
  of perspectives of experienced users with chronic conditions.
\newblock \emph{BMC family practice}, 20\penalty0 (1):\penalty0 57, 2019.

\bibitem[Hinton et~al.(2015)Hinton, Vinyals, and Dean]{hinton2015distilling}
Geoffrey Hinton, Oriol Vinyals, and Jeff Dean.
\newblock Distilling the knowledge in a neural network.
\newblock \emph{arXiv preprint arXiv:1503.02531}, 2015.

\bibitem[Huang et~al.(2019)Huang, Altosaar, and
  Ranganath]{huang2019clinicalbert}
Kexin Huang, Jaan Altosaar, and Rajesh Ranganath.
\newblock Clinicalbert: Modeling clinical notes and predicting hospital
  readmission.
\newblock \emph{arXiv preprint arXiv:1904.05342}, 2019.

\bibitem[Kim(2014)]{kim2014convolutional}
Yoon Kim.
\newblock Convolutional neural networks for sentence classification.
\newblock In \emph{Proceedings of the 2014 Conference on Empirical Methods in
  Natural Language Processing, {EMNLP} 2014, October 25-29, 2014, Doha, Qatar,
  {A} meeting of SIGDAT, a Special Interest Group of the {ACL}}, pages
  1746--1751, 2014.
\newblock URL \url{http://aclweb.org/anthology/D/D14/D14-1181.pdf}.

\bibitem[Kingma and Ba(2015)]{kingma14adam}
Diederik~P. Kingma and Jimmy Ba.
\newblock Adam: {A} method for stochastic optimization.
\newblock In Yoshua Bengio and Yann LeCun, editors, \emph{3rd International
  Conference on Learning Representations, {ICLR} 2015, San Diego, CA, USA, May
  7-9, 2015, Conference Track Proceedings}, 2015.
\newblock URL \url{http://arxiv.org/abs/1412.6980}.

\bibitem[Kiros et~al.(2014)Kiros, Salakhutdinov, and
  Zemel]{kiros2014multimodal}
Ryan Kiros, Ruslan Salakhutdinov, and Rich Zemel.
\newblock Multimodal neural language models.
\newblock In \emph{International conference on machine learning}, pages
  595--603, 2014.

\bibitem[Lee et~al.(2020)Lee, Yoon, Kim, Kim, Kim, So, and
  Kang]{lee2020biobert}
Jinhyuk Lee, Wonjin Yoon, Sungdong Kim, Donghyeon Kim, Sunkyu Kim, Chan~Ho So,
  and Jaewoo Kang.
\newblock Biobert: a pre-trained biomedical language representation model for
  biomedical text mining.
\newblock \emph{Bioinformatics}, 36\penalty0 (4):\penalty0 1234--1240, 2020.

\bibitem[Li et~al.(2015)Li, Liao, Lan, Du, and Yang]{li2015zero}
Xirong Li, Shuai Liao, Weiyu Lan, Xiaoyong Du, and Gang Yang.
\newblock Zero-shot image tagging by hierarchical semantic embedding.
\newblock In \emph{Proceedings of the 38th International ACM SIGIR Conference
  on Research and Development in Information Retrieval}, pages 879--882, 2015.

\bibitem[Ma et~al.(2016)Ma, Cambria, and Gao]{ma2016label}
Yukun Ma, Erik Cambria, and Sa~Gao.
\newblock Label embedding for zero-shot fine-grained named entity typing.
\newblock In \emph{Proceedings of COLING 2016, the 26th International
  Conference on Computational Linguistics: Technical Papers}, pages 171--180,
  2016.

\bibitem[Maaten and Hinton(2008)]{maaten2008visualizing}
Laurens van~der Maaten and Geoffrey Hinton.
\newblock Visualizing data using t-sne.
\newblock \emph{Journal of machine learning research}, 9\penalty0
  (Nov):\penalty0 2579--2605, 2008.

\bibitem[Moosavi-Dezfooli et~al.(2016)Moosavi-Dezfooli, Fawzi, and
  Frossard]{moosavi2016deepfool}
Seyed-Mohsen Moosavi-Dezfooli, Alhussein Fawzi, and Pascal Frossard.
\newblock Deepfool: a simple and accurate method to fool deep neural networks.
\newblock In \emph{Proceedings of the IEEE conference on computer vision and
  pattern recognition}, pages 2574--2582, 2016.

\bibitem[Para et~al.(2019)Para, Sezer, and Sever]{para2019clinical}
Oznur~Esra Para, Ebru~Akcpinar Sezer, and Hayri Sever.
\newblock Clinical decision support systems: From the perspective of small and
  imbalanced data set.
\newblock \emph{Health Informatics Vision: From Data via Information to
  Knowledge}, 262:\penalty0 344, 2019.

\bibitem[Parambath et~al.(2014)Parambath, Usunier, and
  Grandvalet]{parambath2014optimizing}
Shameem~Puthiya Parambath, Nicolas Usunier, and Yves Grandvalet.
\newblock Optimizing f-measures by cost-sensitive classification.
\newblock In \emph{Advances in Neural Information Processing Systems}, pages
  2123--2131, 2014.

\bibitem[Paszke et~al.(2019)Paszke, Gross, Massa, Lerer, Bradbury, Chanan,
  Killeen, Lin, Gimelshein, Antiga, et~al.]{paszke2019pytorch}
Adam Paszke, Sam Gross, Francisco Massa, Adam Lerer, James Bradbury, Gregory
  Chanan, Trevor Killeen, Zeming Lin, Natalia Gimelshein, Luca Antiga, et~al.
\newblock Pytorch: An imperative style, high-performance deep learning library.
\newblock In \emph{Advances in Neural Information Processing Systems}, pages
  8024--8035, 2019.

\bibitem[Pedregosa et~al.(2011)Pedregosa, Varoquaux, Gramfort, Michel, Thirion,
  Grisel, Blondel, Prettenhofer, Weiss, Dubourg, Vanderplas, Passos,
  Cournapeau, Brucher, Perrot, and Duchesnay]{scikit-learn}
F.~Pedregosa, G.~Varoquaux, A.~Gramfort, V.~Michel, B.~Thirion, O.~Grisel,
  M.~Blondel, P.~Prettenhofer, R.~Weiss, V.~Dubourg, J.~Vanderplas, A.~Passos,
  D.~Cournapeau, M.~Brucher, M.~Perrot, and E.~Duchesnay.
\newblock Scikit-learn: Machine learning in {P}ython.
\newblock \emph{Journal of Machine Learning Research}, 12:\penalty0 2825--2830,
  2011.

\bibitem[Pennington et~al.(2014)Pennington, Socher, and
  Manning]{pennington2014glove}
Jeffrey Pennington, Richard Socher, and Christopher~D Manning.
\newblock Glove: Global vectors for word representation.
\newblock In \emph{Proceedings of the 2014 conference on empirical methods in
  natural language processing (EMNLP)}, pages 1532--1543, 2014.

\bibitem[Qian et~al.(2016)Qian, Li, Zhu, Zhou, Luo, and
  Luo]{qian2016speculation}
Zhong Qian, Peifeng Li, Qiaoming Zhu, Guodong Zhou, Zhunchen Luo, and Wei Luo.
\newblock Speculation and negation scope detection via convolutional neural
  networks.
\newblock In \emph{Proceedings of the 2016 Conference on Empirical Methods in
  Natural Language Processing}, pages 815--825, 2016.

\bibitem[Radford et~al.(2019)Radford, Wu, Child, Luan, Amodei, and
  Sutskever]{radford2019language}
Alec Radford, Jeff Wu, Rewon Child, David Luan, Dario Amodei, and Ilya
  Sutskever.
\newblock Language models are unsupervised multitask learners.
\newblock 2019.

\bibitem[Ramsey et~al.(2018)Ramsey, Lanzo, Huston-Paterson, Tomaszewski, and
  Trent]{ramsey2018increasing}
Alexandra Ramsey, Erin Lanzo, Hattie Huston-Paterson, Kathy Tomaszewski, and
  Maria Trent.
\newblock Increasing patient portal usage: preliminary outcomes from the
  mychart genius project.
\newblock \emph{Journal of Adolescent Health}, 62\penalty0 (1):\penalty0
  29--35, 2018.

\bibitem[Rodriguez-Serrano and Perronnin(2015)]{rodriguez2015label}
Jose~Antonio Rodriguez-Serrano and Florent~C Perronnin.
\newblock Label-embedding for text recognition, April~14 2015.
\newblock US Patent 9,008,429.

\bibitem[Salton and Buckley(1988)]{salton1988term}
Gerard Salton and Christopher Buckley.
\newblock Term-weighting approaches in automatic text retrieval.
\newblock \emph{Information processing \& management}, 24\penalty0
  (5):\penalty0 513--523, 1988.

\bibitem[Sanh et~al.(2019)Sanh, Debut, Chaumond, and Wolf]{sanh2019distilbert}
Victor Sanh, Lysandre Debut, Julien Chaumond, and Thomas Wolf.
\newblock Distilbert, a distilled version of bert: smaller, faster, cheaper and
  lighter.
\newblock \emph{arXiv preprint arXiv:1910.01108}, 2019.

\bibitem[Sieck et~al.(2017)Sieck, Hefner, Schnierle, Florian, Agarwal, Rundell,
  and McAlearney]{sieck2017rules}
Cynthia~J Sieck, Jennifer~L Hefner, Jeanette Schnierle, Hannah Florian, Aradhna
  Agarwal, Kristen Rundell, and Ann~Scheck McAlearney.
\newblock The rules of engagement: perspectives on secure messaging from
  experienced ambulatory patient portal users.
\newblock \emph{JMIR medical informatics}, 5\penalty0 (3):\penalty0 e13, 2017.

\bibitem[Strubell et~al.(2019)Strubell, Ganesh, and
  McCallum]{strubell2019energy}
Emma Strubell, Ananya Ganesh, and Andrew McCallum.
\newblock Energy and policy considerations for deep learning in nlp.
\newblock \emph{arXiv preprint arXiv:1906.02243}, 2019.

\bibitem[Sulieman et~al.(2017)Sulieman, Gilmore, French, Cronin, Jackson,
  Russell, and Fabbri]{sulieman2017classifying}
Lina Sulieman, David Gilmore, Christi French, Robert~M Cronin, Gretchen~Purcell
  Jackson, Matthew Russell, and Daniel Fabbri.
\newblock Classifying patient portal messages using convolutional neural
  networks.
\newblock \emph{Journal of biomedical informatics}, 74:\penalty0 59--70, 2017.

\bibitem[Tafti et~al.(2019)Tafti, Fu, Khurana, Mastorakos, Poole, Traub,
  Yiannias, and Liu]{tafti2019artificial}
Ahmad~P Tafti, Sunyang Fu, Aditya Khurana, George~M Mastorakos, Kenneth~G
  Poole, Stephen~J Traub, James~A Yiannias, and Hongfang Liu.
\newblock Artificial intelligence to organize patient portal messages: a
  journey from an ensemble deep learning text classification to rule-based
  named entity recognition.
\newblock In \emph{2019 IEEE International Conference on Bioinformatics and
  Biomedicine (BIBM)}, pages 1380--1387. IEEE, 2019.

\bibitem[Tang et~al.(2019)Tang, Lu, Liu, Mou, Vechtomova, and
  Lin]{tang2019distilling}
Raphael Tang, Yao Lu, Linqing Liu, Lili Mou, Olga Vechtomova, and Jimmy Lin.
\newblock Distilling task-specific knowledge from bert into simple neural
  networks.
\newblock \emph{arXiv preprint arXiv:1903.12136}, 2019.

\bibitem[Tong and Koller(2001)]{tong2001support}
Simon Tong and Daphne Koller.
\newblock Support vector machine active learning with applications to text
  classification.
\newblock \emph{Journal of machine learning research}, 2\penalty0
  (Nov):\penalty0 45--66, 2001.

\bibitem[Vaswani et~al.(2017)Vaswani, Shazeer, Parmar, Uszkoreit, Jones, Gomez,
  Kaiser, and Polosukhin]{vaswani2017attention}
Ashish Vaswani, Noam Shazeer, Niki Parmar, Jakob Uszkoreit, Llion Jones,
  Aidan~N Gomez, {\L}ukasz Kaiser, and Illia Polosukhin.
\newblock Attention is all you need.
\newblock In \emph{Advances in neural information processing systems}, pages
  5998--6008, 2017.

\bibitem[Wang et~al.(2018)Wang, Li, Wang, Zhang, Shen, Zhang, Henao, and
  Carin]{wang2018joint}
Guoyin Wang, Chunyuan Li, Wenlin Wang, Yizhe Zhang, Dinghan Shen, Xinyuan
  Zhang, Ricardo Henao, and Lawrence Carin.
\newblock Joint embedding of words and labels for text classification.
\newblock In \emph{ACL}, 2018.

\bibitem[Wu et~al.(2016)Wu, Schuster, Chen, Le, Norouzi, Macherey, Krikun, Cao,
  Gao, Macherey, et~al.]{wu2016google}
Yonghui Wu, Mike Schuster, Zhifeng Chen, Quoc~V Le, Mohammad Norouzi, Wolfgang
  Macherey, Maxim Krikun, Yuan Cao, Qin Gao, Klaus Macherey, et~al.
\newblock Google's neural machine translation system: Bridging the gap between
  human and machine translation.
\newblock \emph{arXiv preprint arXiv:1609.08144}, 2016.

\bibitem[Yang et~al.(2016)Yang, Yang, Dyer, He, Smola, and
  Hovy]{yang2016hierarchical}
Zichao Yang, Diyi Yang, Chris Dyer, Xiaodong He, Alex Smola, and Eduard Hovy.
\newblock Hierarchical attention networks for document classification.
\newblock In \emph{Proceedings of the 2016 conference of the North American
  chapter of the association for computational linguistics: human language
  technologies}, pages 1480--1489, 2016.

\bibitem[Zhang et~al.(2017)Zhang, Xiao, Chen, Wang, and Jin]{zhang2017multi}
Honglun Zhang, Liqiang Xiao, Wenqing Chen, Yongkun Wang, and Yaohui Jin.
\newblock Multi-task label embedding for text classification.
\newblock \emph{arXiv preprint arXiv:1710.07210}, 2017.

\bibitem[Zhang et~al.(2008)Zhang, Yoshida, and Tang]{zhang2008tfidf}
Wen Zhang, Taketoshi Yoshida, and Xijin Tang.
\newblock Tfidf, lsi and multi-word in information retrieval and text
  categorization.
\newblock In \emph{2008 IEEE International Conference on Systems, Man and
  Cybernetics}, pages 108--113. IEEE, 2008.

\bibitem[Zhao et~al.(2018)Zhao, Wong, and Tsui]{zhao2018framework}
Yang Zhao, Zoie Shui-Yee Wong, and Kwok~Leung Tsui.
\newblock A framework of rebalancing imbalanced healthcare data for rare
  events’ classification: a case of look-alike sound-alike mix-up incident
  detection.
\newblock \emph{Journal of healthcare engineering}, 2018, 2018.

\bibitem[Zhu et~al.(2015)Zhu, Kiros, Zemel, Salakhutdinov, Urtasun, Torralba,
  and Fidler]{zhu2015aligning}
Yukun Zhu, Ryan Kiros, Rich Zemel, Ruslan Salakhutdinov, Raquel Urtasun,
  Antonio Torralba, and Sanja Fidler.
\newblock Aligning books and movies: Towards story-like visual explanations by
  watching movies and reading books.
\newblock In \emph{Proceedings of the IEEE international conference on computer
  vision}, pages 19--27, 2015.

\end{thebibliography}

%

\end{document}